\documentclass{article}
\usepackage{ijcai25}

\usepackage{times}
\usepackage{soul}
\usepackage{url}
\usepackage[hidelinks]{hyperref}
\usepackage[utf8]{inputenc}
\usepackage[small]{caption}
\usepackage{graphicx}
\usepackage{amsmath}
\usepackage{amsthm}
\usepackage{booktabs}
\usepackage{algorithm}
\usepackage{algorithmic}
\usepackage[switch]{lineno}
\usepackage{enumitem}
\usepackage{caption} 
\usepackage{multirow}
\usepackage{hyperref} 
\usepackage{subfig}

\title{Single-Node Trigger Backdoor Attacks in Graph-Based Recommendation Systems}

\author{
    Runze Li$^{1,2}$\and Di Jin$^1$\and Xiaobao Wang$^{1,2,}$\footnote{Corresponding author} \and Dongxiao He$^1$\and Bingdao Feng$^1$\And Zhen Wang$^3$\\
    \affiliations
    $^1$College of Intelligence and Computing, Tianjin University, Tianjin, China\\
    $^2$Guangdong Laboratory of Artificial Intelligence and Digital Economy(SZ), Shenzhen, China\\
    $^3$School of Cybersecurity, Northwestern Polytechnical University, Xi'an, China\\
    \emails
    \{lirunze, jindi, wangxiaobao, hedongxiao, fengbingdao\}@tju.edu.cn,
    w-zhen@nwpu.edu.cn
}

\begin{document}

\maketitle

\begin{abstract}
    Graph recommendation systems have been widely studied due to their ability to effectively capture the complex interactions between users and items. However, these systems also exhibit certain vulnerabilities when faced with attacks. The prevailing shilling attack methods typically manipulate recommendation results by injecting a large number of fake nodes and edges. However, such attack strategies face two primary challenges: low stealth and high destructiveness. To address these challenges, this paper proposes a novel graph backdoor attack method that aims to enhance the exposure of target items to the target user in a covert manner, without affecting other unrelated nodes. Specifically, we design a single-node trigger generator, which can effectively expose multiple target items to the target user by inserting only one fake user node. Additionally, we introduce constraint conditions between the target nodes and irrelevant nodes to mitigate the impact of fake nodes on the recommendation system's performance. Experimental results show that the exposure of the target items reaches no less than $50\%$ in $99\%$ of the target users, while the impact on the recommendation system's performance is controlled within approximately $5\%$. 
\end{abstract}

\section{INTRODUCTION}

\begin{figure}[tbp]
    \centering
    \setcounter{subfigure}{0}
    \subfloat[Yelp]{
    \includegraphics[width=0.48\linewidth]{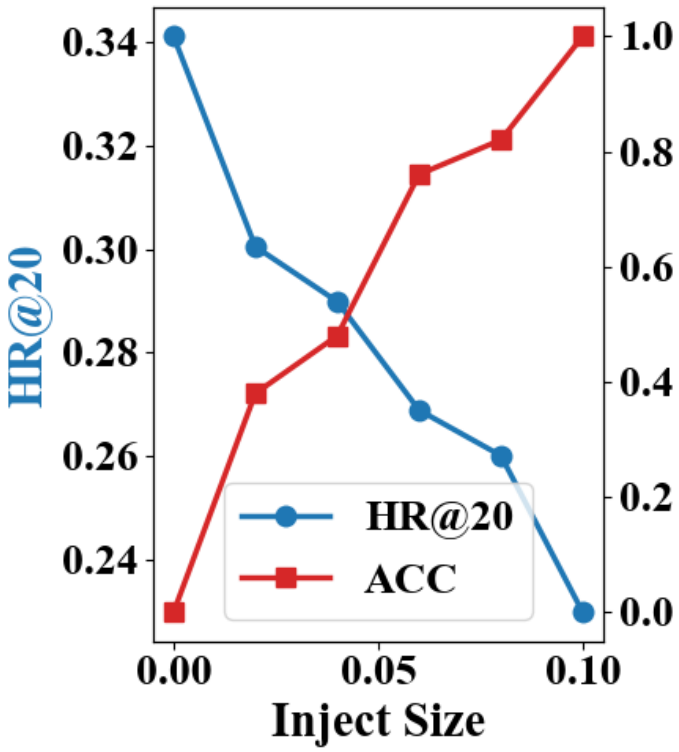}}
    \hspace{0cm}
    \subfloat[Gowalla]{
    \includegraphics[width=0.48\linewidth]{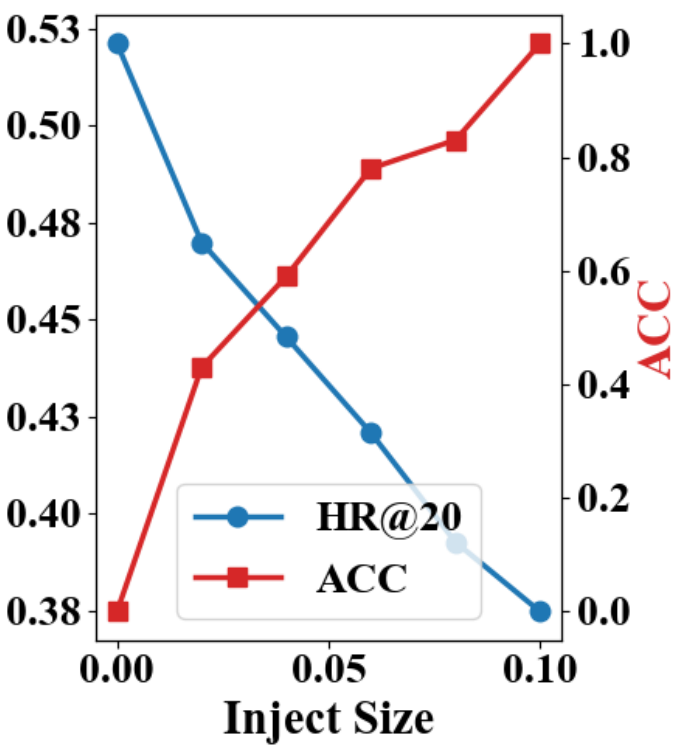}}
    \vspace{0cm} 
    \caption{We investigate the impact of the injected false user ratio on the accuracy of recommendation systems using the AutoAttack method on the Yelp and Gowalla datasets. The 'Inject size' denotes the ratio of injected false users to target users, 'ACC' indicates the proportion of successful recommendations among target users, and 'HR@20' is the evaluation metric for recommendation performance.}
    \label{fig1}
\end{figure}

Recommendation systems play a vital role in modern information societies and are widely applied in areas such as e-commerce, social media, and content platforms. By providing users with personalized content and services, they significantly enhance user experience and platform revenue. Traditional recommendation systems, typically based on collaborative filtering \cite{filter1,filter2}, matrix factorization \cite{matrix1,matrix2}, or sequence model techniques \cite{seq1,seq2}, often struggle to fully explore the potential high-order relationships and contextual information between users and items. To better capture complex interaction patterns, Graph Neural Networks (GNNs) have been widely adopted in recommendation systems due to their ability to model relational data\cite{w1,w2,w3,w4,w5}. By leveraging GNNs,Graph-based Recommendation Systems (GRSs) can effectively represent users, items, and their interactions as a graph structure, enabling a more comprehensive discovery of latent user behavior patterns.By modeling high-order connectivity and complex relationships, GRSs improve personalized recommendation accuracy \cite{Yang2018GRS6,Dong2024GRS1,Luzhi}.

However, recommendation systems based on Graph Neural Networks have certain vulnerabilities \cite{Vulnerability,Robust,w6}, and recent studies have focused on shilling attacks in recommendation systems. Attackers can achieve targeted attacks by carefully designing perturbations to nodes or edges, such as GOAT \cite{GSA1-2021} who first proposed injecting fake user features and link structures in the graph to recommend target items to the user group. AutoAttack \cite{Autoattack} consider enhancing the exposure of target items within the interested group from a more realistic perspective. However, current methods have two significant flaws, as shown in Figure \ref{fig1}. First, to achieve better attack effectiveness, existing methods require injecting a large number of fake nodes, which lack stealth. Second, as graph-based recommendation systems rely on the bipartite graph structure between users and items to generate personalized recommendations through information propagation via historical interaction edges, the large amount of false information injected by shilling attacks can corrupt the graph structure and the information propagation process, significantly disrupting the accuracy and stability of recommendations.

\begin{figure}[tbp] 
    \centering
    \includegraphics[width=\linewidth]{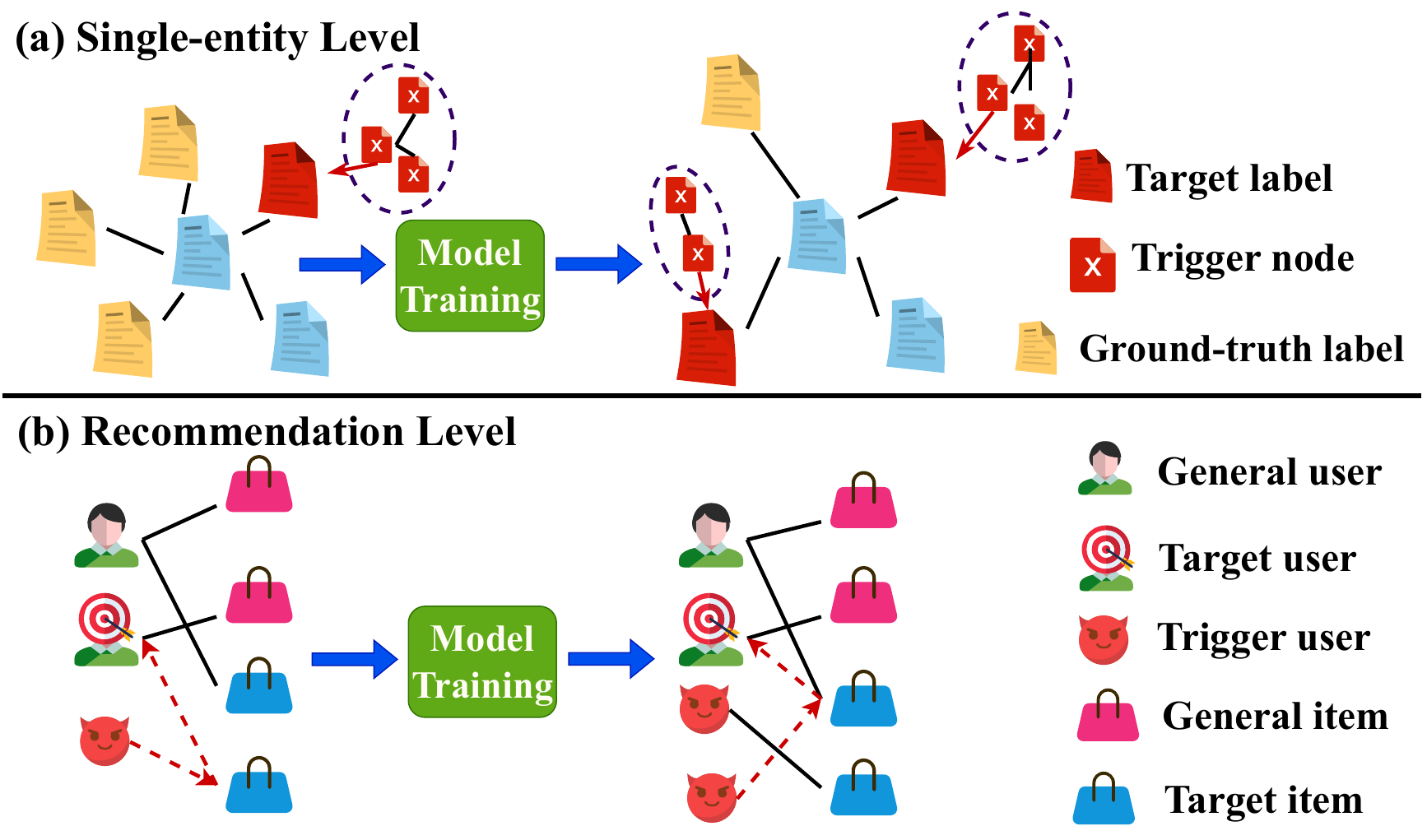} 
    \caption{The difference between (a) traditional single-entity backdoor attacks and (b) recommendation system backdoor attacks.} 
    \label{fig2} 
\end{figure}

Graph backdoor attacks are a covert form of graph structure attack, where an attacker quietly implants malicious nodes or edges in the graph to influence the overall behavior of the graph when specific conditions are triggered. The primary advantage of such attacks lies in their stealthiness, as the malicious modifications are typically difficult to detect, allowing the attack to remain dormant and effective without alerting the system \cite{b1,b2,GB,R-bda}. However, directly applying backdoor attacks to recommendation systems presents two challenges. First, current graph backdoor attacks mainly focus on single-entity classification tasks. Specifically, they generate specific triggers for a single node \cite{UGBA,BD-backdoor,b4,b5,b6} or a single graph \cite{GTA,GTA2,b3} to influence its classification results, as shown in Figure \ref{fig2}(a). However, in a graph-based recommendation system, recommendations are generated by computing the similarity between users and items. Therefore, the key challenge is how to generate triggers that can simultaneously affect both target users and items, as shown in Figure \ref{fig2}(b). Second, even if a backdoor attack method is successfully applied to the recommendation system, generating specific triggers for each target item poses a challenge. Once a large number of trigger nodes are injected, the fake information contained in these triggers will still be spread to unrelated users and items through message propagation, significantly disrupting the recommendation performance of the original system. Thus, another key challenge is how to prevent the trigger information from affecting irrelevant nodes.

To address these challenges, we propose a backdoor attack method  named \textbf{S}ingle-\textbf{N}ode \textbf{T}rigger \textbf{B}ackdoor \textbf{A}ttacks in Graph-Based Recommendation Systems(SNT-BA). Specifically, we introduce a single-node trigger generator for target items and users. By inserting a fake user as a trigger into all target items, we can significantly increase the exposure of these items to the target user. We achieve this by introducing latent edges in the graph, allowing the trigger to simultaneously affect both the target items and the users. Moreover, since it is a single-trigger structure, only a single piece of fake information is propagated to irrelevant items and users, thus minimizing the impact on the original recommendation system's performance. In addition, we design constraints between user and item nodes to ensure the stability of the distribution of user and item nodes, further enhancing the stealth of the attack. Finally, our proposed attack method is an end-to-end approach, where, during application, inserting the trigger user into the target items is sufficient to achieve the attack objective. The contributions of our work are as follows:

\begin{itemize}[left=0em]
    \item We propose a backdoor attack approach for graph-based recommendation systems, which, to the best of our knowledge, is the first work to apply backdoor attacks in the context of graph recommendation systems.
    \item We propose a more effective recommendation attack method, which not only solves the problems of low stealthiness and high destructiveness in existing graph recommendation attack methods, but also reduces the attack cost.
    \item We conduct experiments on multiple datasets, ensuring that the exposure of the target item to at least $50\%$ of the target users is achieved in $99\%$ of cases, while the impact on the performance of the recommendation system is controlled to around $5\%$.
\end{itemize}

\section{PRELIMINARY}

In this section, we will define graph-based recommendation systems and backdoor attack methods targeting graph neural networks.

\subsection{Graph-based Recommendation System}
A graph-based recommendation system models users and items as nodes, with their interactions as edges. It leverages the graph structure to infer user preferences for unseen items based on connectivity patterns.

Let the graph in the recommendation system be defined as $G = (V, E)$, where $V$ is the set of nodes, $V = U \cup I$, $U$ is the set of user nodes, and $I$ is the set of item nodes. $E$ is the set of edges that represents the interactions between users and items. Each edge $e_{u,i} \in E$ represents an interaction between user $u \in U$ and item $i \in I$.

In graph-based recommendation systems, the goal of the model is to learn low-dimensional vector representations for users and items, and to perform prediction tasks based on these representations. Let $h_u(\theta)$ represent the embedding vector of user $u$, and $h_i(\theta)$ represent the embedding vector of item $i$, where $\theta$ denotes the model parameters. These embedding vectors are optimized by learning the graph structure to improve the accuracy of the recommendation.

A typical objective in the recommendation task is to predict the rating that a user would give to an item. Suppose the goal is to predict the rating $\hat{r}_{ui}$ that user $u$ gives to item $i$, this rating can be computed using the following formula:
 \begin{equation} \label{eq1}
    \hat{r}_{ui} = f(h_u(\theta), h_i(\theta)),
\end{equation}%
where $f(\cdot)$ is the rating prediction function. By training the model and optimizing the parameters $\theta$, the prediction error between the predicted and actual ratings is minimized, thereby accomplishing the recommendation task.

\subsection{Graph Backdoor Attacks on GRSs}
In recommendation systems, the goal of a backdoor attack is to manipulate the similarity prediction between user $u$ and item $i$ by introducing triggers, such as fake users or false interactions.

Specifically, after training the model with a backdoor attack, we leave a backdoor in the recommendation graph. During the application of the model, once the trigger is inserted for the target item, the model will output a higher similarity score for the target user and target item, thereby prioritizing the items that the attacker wants to be recommended. In contrast, without the trigger, the model will perform the recommendation task normally, without any detectable anomalies. In a backdoor attack, the similarity prediction can be represented as:
\begin{equation}
\hat{r}_{ui} =
\begin{cases} 
\max\limits_{G \in \mathcal{G}_{\text{trigger}}} f(h_u(\theta), h_{i^*}(\theta)), & \text{if } G \text{ contains the trigger}, \\
f(h_u(\theta), h_i(\theta)), & \text{otherwise}.
\end{cases}
\end{equation}
Let $\mathcal{G}_{\text{trigger}}$ be the attacked graph with the inserted trigger. When the trigger is present, our goal is to maximize $\hat{r}_{ui^*}$, where $i^*$ is the target item, to ensure that the target item is prioritized in the recommendation. At the same time, when no trigger is injected, the model continues to make recommendations according to Equation (\ref{eq1}), ensuring the stealthiness of the model.

\section{METHODS}
Previous graph recommendation system attack methods have suffered from poor stealth and strong disruption due to the injection of a large number of fake nodes. In this paper, we propose a more covert and flexible backdoor attack method. This method can dynamically generate a single specific trigger based on the features of all attacked nodes, thereby avoiding the limitations of attack node structure and quantity on the attack's effectiveness. This section will elaborate on the structure and principles of the proposed model.

\subsection{The Backdoor Attack Framework}
\begin{figure*}[htbp]
    \centering
    \includegraphics[width=\textwidth]{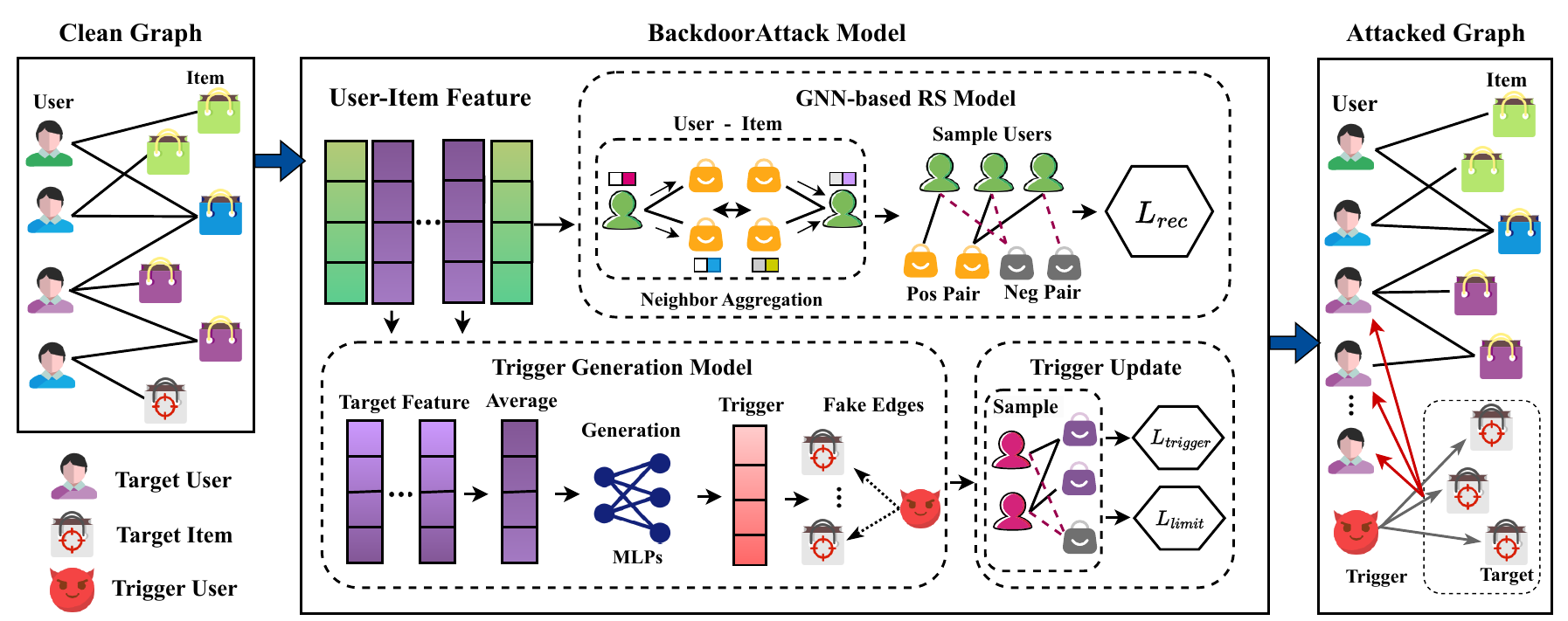}
    \caption{The framework of SNT-BA consists of two modules: the Trigger Generator and the Recommendation System Training Module. By designing triggers, the target items can significantly improve their rankings after the insertion of the trigger. At the same time, the optimization of the constraint loss function ensures the stability of recommendation performance for unrelated users and items.}
    \label{fig3}
\end{figure*}

Figure \ref{fig3} shows the architecture of the backdoor attack model for graph-based recommendation systems. Our task is to take a clean input graph and, through the designed backdoor attack model, enable any target item in the attacked graph to significantly improve its ranking in the target user's candidate item list after the trigger is inserted. At the same time, it is necessary to ensure that, prior to the attack, the contaminated model remains stealthy. Furthermore, after inserting the trigger, the model's recommendation performance for irrelevant users and items should remain stable. To achieve this goal, we divide the backdoor attack model into two modules: the trigger generation training module and the recommendation system training module. The trigger generation module creates specific triggers for the target items according to their features, ensuring that the insertion of these triggers will boost the ranking of the target item in the candidate set. The recommendation training module is responsible for ensuring that the attacked graph representation, when used for recommendations, maintains the accuracy of the model.
\subsection{Trigger Generation}
\noindent\textbf{Fake User Generation: }  
The objective of the backdoor attack is to recommend the target item to the target class of users $t \in T$ after injecting the trigger. To ensure the stealthiness of the trigger node, we only inject a single fake user node $u^*$ into the recommendation system. To ensure that the fake user node only influences the target item, for all target items $s \in S$, we average their representations $f_{t} \in R^{|S| \times d}$ to find the centroid node feature $f_{s^*} \in R^{1 \times d}$, where $d$ is the dimension of the node features. For the fake user node's feature $f_{u^*}$, we generate its feature representation based on $f_{s^*}$ using a two-layer neural network, as follows:
\begin{equation}
f_{u^*} = \sigma(f_{s^*}W_1+b_1)W_2+b_2.
\end{equation}
To ensure that the target item is preferentially recommended to the target class of users $U_t$, it is necessary to guarantee that, after injecting the trigger, the similarity between the target item $S$ and the users of the target class $T$ is higher than that of other items in the same class. Therefore, we select a number of negative sample items from the non-target items in the target class, equal in quantity to the target items. We design the loss function $L_{trigger}$ to ensure that the target user should be more similar to the target item in this round of recommendations. The loss function is designed as follows:
\begin{equation}
L_{\text{trigger}}(\mathcal{D} | \Theta) = \sum_{t \in T, s \in S,p \in P} -\ln \sigma(\hat{r}_{ts} - \hat{r}_{tp}) + \lambda_\Theta \|\Theta\|^2,
\end{equation}
where $\mathcal{D}$ represents all recommendation data, $\Theta$ represents the parameters of the recommendation model, $P$ is the set of candidate target items ($S \in P$), and $p \in P$ refers to the candidate item selected as a negative sample in the current iteration. $\hat{r}_{ts}$ represents the rating of the target item by the target user, $\hat{r}_{tp}$ represents the rating of candidate negative items by the target user, and $\lambda_\Theta \|\Theta\|^2$ is the regularization coefficient used to control the complexity of the parameters.

\noindent\textbf{Constraints on Attacked Nodes: }  
Since our task involves reserving a backdoor during the training process, and each round of trigger injection simultaneously affects both the target user $T$ and the target item $S$, this will lead to a shift in the position of candidate items and the target user in the original feature space after multiple training rounds. Specifically, because items of the same type are competing, the candidate items will first be pushed away from the target user, and only after the injection of triggers will the representations of the target items and the target user be pulled closer together. Since the target user will also interact with noncandidate items $N$, this will reduce the exposure of the intended target items, which could be detected by the recommendation system.

Therefore, we need to design a constraint loss to control the relative positions of users and items in the original feature space. In each round, we randomly sample the same number of interactions between the target user $T$ and other noncandidate items $N$ as negative samples for interactions with the candidate item $P$, thereby ensuring the relative stability of the distributions of different types of nodes. The specific constraint loss is as follows:
\begin{equation}
\label{lcons}
L_{\text{limit}}(\mathcal{D} | \Theta) = \sum_{(t, p, n) \in \mathcal{D}} -\ln \sigma(\hat{r}_{tp} - \hat{r}_{tn}) + \lambda_\Theta \|\Theta\|^2.
\end{equation}

\subsection{GNN-based Recommendation}
Since backdoor attacks do not alter the original model's training procedure, we directly use LightGCN \cite{LightGCN} as the surrogate model for the attack. The core idea of LightGCN is to learn node representations through neighborhood-based information propagation on the graph structure. Unlike traditional GCNs, LightGCN simplifies the convolution operation by disregarding the node feature matrix and instead propagating information solely using the graph adjacency matrix. Specifically, this can be expressed as follows.
\begin{equation}
\mathbf{h}_v^{(k)} = \sum_{u \in \mathcal{N}(v)} \frac{1}{\sqrt{d_v d_u}} \mathbf{h}_u^{(k-1)}.
\end{equation}
In this context, $\mathbf{h}_v^{(k)}$ represents the embedding of node $v$ at the $k$-th layer, $\mathcal{N}(v) $ denotes the set of neighbors of node $v$, $d_v$ is the degree of node $v$, and $d_u$ is the degree of node $u$.

In LightGCN, the convolution operation at each layer involves a simple weighted summation, so the multi-layer graph convolution operation is essentially multiple weighted averages of different neighbor information. The final node representation $\mathbf{h}_v^{(k)}$ is the result after $k$ layers of convolution. Specifically, the final representation of node $v$ can be expressed as:
\begin{equation}
\mathbf{h}_v = \sum_{k=0}^{K-1} \mathbf{h}_v^{(k)},
\end{equation}
where $K$ is the number of graph convolution layers, representing the node's embedding after multiple rounds of neighbor information propagation.

During training, the BPR loss function is typically used to optimize the model parameters. The BPR loss function optimizes by maximizing the user's preference for positive items relative to negative items. The BPR loss function is defined as:

\begin{equation}
L_{\text{rec}}(\mathcal{D} | \Theta) = \sum_{u,i,j} -\ln \sigma(\hat{r}_{ui} - \hat{r}_{uj}) + \lambda_\Theta \|\Theta\|^2,
\end{equation}
where $(u,i,j)$ represents a triplet, with $i$ being the positive item interacted with by user $u$, $j$ being the negative item, and $\sigma$ denoting the sigmoid function.

\subsection{Joint Optimization Function}
The trigger generator model and the recommendation model are trained simultaneously, ensuring that the recommendation performance for other types of nodes remains intact after trigger injection. Considering all the aforementioned loss conditions, we adopt a joint optimization loss function, which is defined as follows:
\begin{equation}
L = \alpha L_{\text{trigger}} + \beta L_{\text{limit}} + \gamma L_{\text{rec}},
\end{equation}
where $\alpha$, $\beta$ and $\gamma$ are hyper-parameters that control the impact of each optimization objective on the final attack effects.

\section{EXPERIMENTS}

In this section, we will evaluate our method on several real-world recommendation datasets to address the following research questions:
\begin{itemize}[left=0em]
    \item \textbf{RQ1}: Can the proposed method effectively perform backdoor attacks on graph-based recommendation systems, and can it increase the appearance of target items in the recommendation lists of more target users?
    \item  \textbf{RQ2}: How does the constraint loss function enhance the stealthiness of the attack, and how can it ensure the attack remains undetected as much as possible before execution?
    \item  \textbf{RQ3}: Why is the use of multiple triggers not preferred for backdoor attacks in graph-based recommendation systems? What are the advantages of using a single trigger compared to multiple structural triggers?
\end{itemize}

\subsection{Experimental Settings}

\noindent\textbf{Datasets:}
We conduct experiments on four real-world datasets, namely Gowalla, Amazon, Yelp and MovieLens. When constructing the user-item interaction graph, we hypothesize that if a user's rating for an item exceeds the average rating, it can be determined that the user has an affinity for the item, and an interaction edge between the user and the item is established.The dataset statistics are shown in Table \ref{table:dataset}.

\begin{itemize}[left=0em]
    \item \textbf{Gowalla}\footnote{\url{http://snap.stanford.edu/data/loc-gowalla.html}}: is a popular check-in dataset that contains data on users' check-in times, locations and social relationships.
    \item  \textbf{Amazon}\footnote{\url{https://snap.stanford.edu/data/amazon/}}: is a collection of user-item interactions from Amazon, specifically focusing on books. It includes data such as user reviews, ratings, and metadata about books.
    \item  \textbf{Yelp}\footnote{\url{https://www.kaggle.com/yelp-dataset/yelp-dataset}}: is a publicly available collection of data from Yelp, a popular platform for user reviews of local businesses such as restaurants, cafes, shops, and service providers.
    \item  \textbf{MovieLens}\footnote{\url{https://www.kaggle.com/datasets/movielens-100k-dataset}}: is a widely used dataset for recommendation system research and experiments, containing user ratings for movies and metadata information about the movies.
\end{itemize}

\begin{table}[t]
\centering
\renewcommand{\arraystretch}{1.2} 
\begin{tabular}{lccc}
\toprule
\textbf{Datasets} & \textbf{\#Users} & \textbf{\#Items} & \textbf{\#Interactions} \\
\hline
Gowalla & 29,858 & 40,981 & 1,027,370 \\
Amazon & 52,643 & 91,599 & 2,984,108 \\
Yelp & 31,831 & 40,841 & 1,666,869 \\
MovieLens & 20,982 & 16,482 & 454,011 \\
\bottomrule
\end{tabular}
\caption{The statistics of datasets.}
\label{table:dataset}
\end{table}

\begin{table*}[h!]
    \centering
    \renewcommand{\arraystretch}{1.5}  
    \begin{tabular}{c c c c c c c c c c c c c}
        \toprule
        \multirow{2}{*}{\textbf{Attacks}} & \multicolumn{3}{c}{\textbf{Yelp}} & \multicolumn{3}{c}{\textbf{Gowalla}} & \multicolumn{3}{c}{\textbf{Amazon}} & \multicolumn{3}{c}{\textbf{Movielens}} \\ \cmidrule(lr){2-4} \cmidrule(lr){5-7} \cmidrule(lr){8-10} \cmidrule(lr){11-13}
                                              & ACC & CVR & HR@20 & ACC & CVR & HR@20 & ACC & CVR & HR@20 & ACC & CVR & HR@20 \\ \midrule
        Random                                &0.22     &0.15     &0.2238       &0.27    &0.21     &0.3587        &0.24     &0.17     &0.1248       & 0.19    & 0.14    & 0.4715    \\ 
        Popular                               &0.30     &0.25     & 0.2137      &0.35     &0.30     & 0.3373      &  0.32    & 0.28    & 0.1065      & 0.26    & 0.22    & 0.4454    \\ 
        Vote                                  & 0.37    & 0.32   & 0.2251      &0.42     &0.53     & 0.3497       &0.49     & 0.45    &  0.1249     & 0.43    & 0.38    & 0.4787     \\ 
        GSPAttack                            & 0.68    & 0.53    & 0.2354      & 0.75    & 0.62    & 0.3412       & 0.63    & 0.55     & 0.1218      & 0.68    & 0.51    & 0.4962     \\ 
        AutoAttack                            & 0.81    & 0.56    &0.2525       & 0.87    & 0.77    & 0.3715       & 0.82    & 0.64     & 0.1384      & 0.85    & 0.65    & 0.5223     \\ \hline
        \textbf{Ours }                                 &  \textbf{1}   &  \textbf{0.71}   & \textbf{0.3245 }     & \textbf{0.99}    &\textbf{0.92}    & \textbf{0.4997}     & \textbf{0.99}    & \textbf{0.87}    & \textbf{0.1709}      & \textbf{1}    & \textbf{0.89}    & \textbf{0.7324}    \\ 
        \bottomrule
    \end{tabular}
    \caption{Comparison of methods on Yelp, Gowalla, Amazon, and Movielens datasets.}
    \label{tab:results}
\end{table*}

\noindent\textbf{Baselines: } 
We compare our method with several traditional attack and injection attack methods.

\begin{itemize}[left=0em]
    \item \textbf{Random Attack} \cite{random}: The attacker randomly generates a large number of fake users connected to the target items, thereby increasing the degree of the target items and enhancing their exposure. In this paper, we introduce an attack by injecting false users constituting $20\%$ of the original user base.
    \item  \textbf{Popular Attack} \cite{popular}: The attacker generates fake users connected to the target items based on the central representation of each user category. In this paper, $5\%$ of the users from each category are added as fake users for the attack.
    \item  \textbf{Vote Attack}: The attacker generates representations for the most active target users and connects them to the target items. In this paper, $10\% $ of the target class users are added as fake users for the attack.
    \item  \textbf{GSPAttack} \cite{GSPAttack}: The attacker uses a GAN network to learn and generate fake user representations and their connection relationships, thereby achieving the attack effect.
    \item  \textbf{AutoAttack} \cite{Autoattack}: The attacker generates fake user representations based on the target user representations and learns the connection relationships. In this paper, $10\%$ of fake users are generated for the attack.
\end{itemize}

\noindent\textbf{Evaluation Metrics: } 
To quantitatively evaluate the impact of the backdoor attack, we design three metrics to assess both the attack performance and recommendation performance. The success rate and coverage rate are used to evaluate the impact of the backdoor attack, while the hit rate is used to assess the recommendation performance.

\begin{itemize}[left=0em]
    \item \textbf{Access Rate}: The access rate (ACC) measures the proportion of target users successfully attacked, defined as the ratio of the number of target users successfully attacked to the total number of target users:
    \begin{equation}
\text{ACC} = \frac{|\mathcal{T}_S|}{|\mathcal{T}_U|},
\end{equation}
    where $\mathcal{T}_S$ is the set of target users successfully attacked, and  $\mathcal{T}_U$ is the set of all target users.
    \item  \textbf{Coverage Rate}: The coverage rate (CVR) measures the proportion of target items in the recommendation list of target users. For each target user, the coverage rate is the proportion of target items appearing in the top $K$ recommended items. The overall coverage rate is the average of these individual coverage rates:
    \begin{equation}
\text{CVR} = \frac{1}{|\mathcal{T}_U|} \sum_{u \in \mathcal{T}_U} \frac{|\mathcal{T}_P(u) \cap \mathcal{R}_u^{(K)}|}{K},
\end{equation}
where $\mathcal{T}_U$ is the set of target users, $\mathcal{T}_P(u)$ is the set of target items for user $u$, and $\mathcal{R}_u^{(K)}$ is the top $K$ recommended item list for user $u$.In this paper, the value of $K$ is set to $10$.
    \item  \textbf{HR@N}: This metric represents the number of items in the top $N$ items of the recommendation list that have been actually interacted with by the user. It is used to calculate the accuracy of the recommendation system in real-world applications. Specifically, it is expressed as:
    \begin{equation}
\text{HR@N} = \frac{|\{ i \in R_u^{(N)} \mid i \in I_u \}|}{N},
\end{equation}
where $R_u^{(N)}$ is the set of the top $N$ recommended items for user $u$, $ I_u$ is the set of items that user $u$ has actually interacted with, and $ |\{ i \in R_u^{(N)} \mid i \in I_u \}| $ represents the number of items in the top $N$ recommended list that the user has actually interacted with. In this paper, the value of $N$ is set to $20$.
\end{itemize}

\noindent\textbf{Parameter Settings: } 
To evaluate the recommendation performance, we split each dataset into two parts: $80\%$ for training and $20\%$ for testing. The number of target items is set to $20$. The training epoch, node embedding size, and learning rate are set to $1000$, $64$, and $0.001$, respectively.

\subsection{Main Result}
To answer \textbf{RQ1}, we evaluate our method from two aspects: the effectiveness of the backdoor attack and its impact on the original recommendation system. For the backdoor attack effectiveness, we assess the success rate for target users and the coverage of target items appearing in the recommendation lists of target users. To evaluate the impact on recommendation performance, we use the $\text{HR@20}$ metric. The experimental results are shown in Table \ref{tab:results}. We will analyze the performance of each metric in detail to verify the effectiveness of our method.

\begin{itemize}[left=0em]
    \item \textbf{Access Rate}: We demonstrate the occurrence of target items in the top $10$ recommendation lists of target users. The results show that for each dataset, the proposed framework successfully includes the target items in the top $10$ recommendations for no less than $99\%$ of the target users, demonstrating a significant performance improvement. Compared to traditional shilling attack methods, our approach performs more favorably in both attack success rate and recommendation quality, effectively boosting the ranking of target items, thereby achieving a stronger backdoor attack effect.
    \item \textbf{Coverage Rate}: We evaluate the coverage of target items in the top $10$ recommended items for all target users using this metric. The experimental results show that, with our approach, the average coverage of target items in the top $10$ recommended items for all target users exceeds $80\%$. This result indicates that our framework ensures that most target users see the target items in the recommendation list during the attack process, further validating the effectiveness and robustness of our method.
    \item \textbf{Hit Rate}: The experimental results show that, after applying our proposed framework, the hit rate of the original recommendation system for the top $20$ recommended items decreases by no more than $5\%$. This result indicates that our attack method not only successfully introduces the target items into the recommendation lists of target users but also keeps the impact on the overall recommendation quality and user experience at a low level, ensuring both the stealthiness and effectiveness of the attack.

\end{itemize}

\subsection{Experimental Analysis}
In this section, we conduct an in-depth analysis of our approach to validate the stealthiness and low destructiveness of the proposed attack method.

\noindent\textbf{Stealthiness:}
To answer \textbf{RQ2}, We first examine the recommendation performance of the model before launching the attack and compare it with the recommendation performance of an unaffected recommendation system, as shown in Table \ref{tab:3}. Specifically, we compare the recommendation accuracy of the model before the backdoor attack with the recommendation accuracy of the original recommendation system, and calculate the relative decline in recommendation performance of our model compared to the attack-free system.

\begin{table}[b]
    \centering
    \renewcommand{\arraystretch}{1.5}  
    \resizebox{0.48\textwidth}{!}{  
    \begin{tabular}{c c c c c}
    \toprule
    \textbf{Datasets} & \textbf{Yelp} & \textbf{Gowalla} & \textbf{Amazon} & \textbf{Movielens} \\ \hline
    \textbf{Raw} & 0.3412 & 0.5322 & 0.2013 & 0.7605 \\
    \textbf{Attacked} & 0.3372 & 0.5157 & 0.1854 & 0.7484 \\
    \hline
    \textbf{Decline Rate} & 0.0394 & 0.0165 & 0.0789 & 0.0121 \\
    \bottomrule
    \end{tabular}
    }
    \caption{Comparison of datasets: Raw, Attacked, and Decline Rate.}
    \label{tab:3}
\end{table}

The results indicate that our recommendation model performs nearly identically to the original recommendation model, with no significant impact on the user's recommendation experience. Additionally, since the model is end-to-end trained, this effect remains undetected.

\begin{figure}[tbp]
    \centering
    \setcounter{subfigure}{0}
    \subfloat[Feature distribution without the use of constraint loss function]{
    \includegraphics[width=0.48\linewidth]{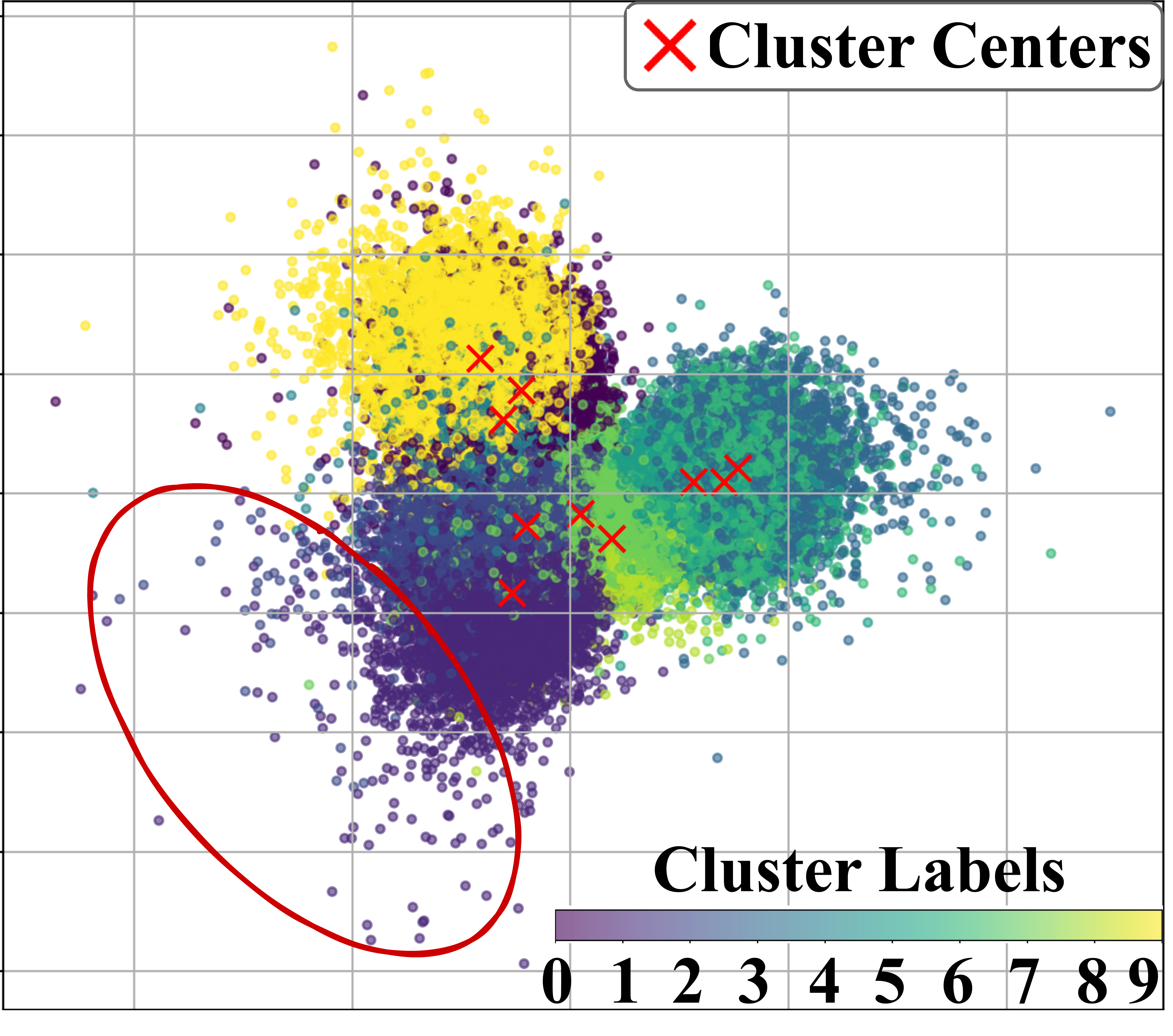}}
    \hspace{0cm}
    \subfloat[Feature distribution with the use of constraint loss function]{
    \includegraphics[width=0.48\linewidth]{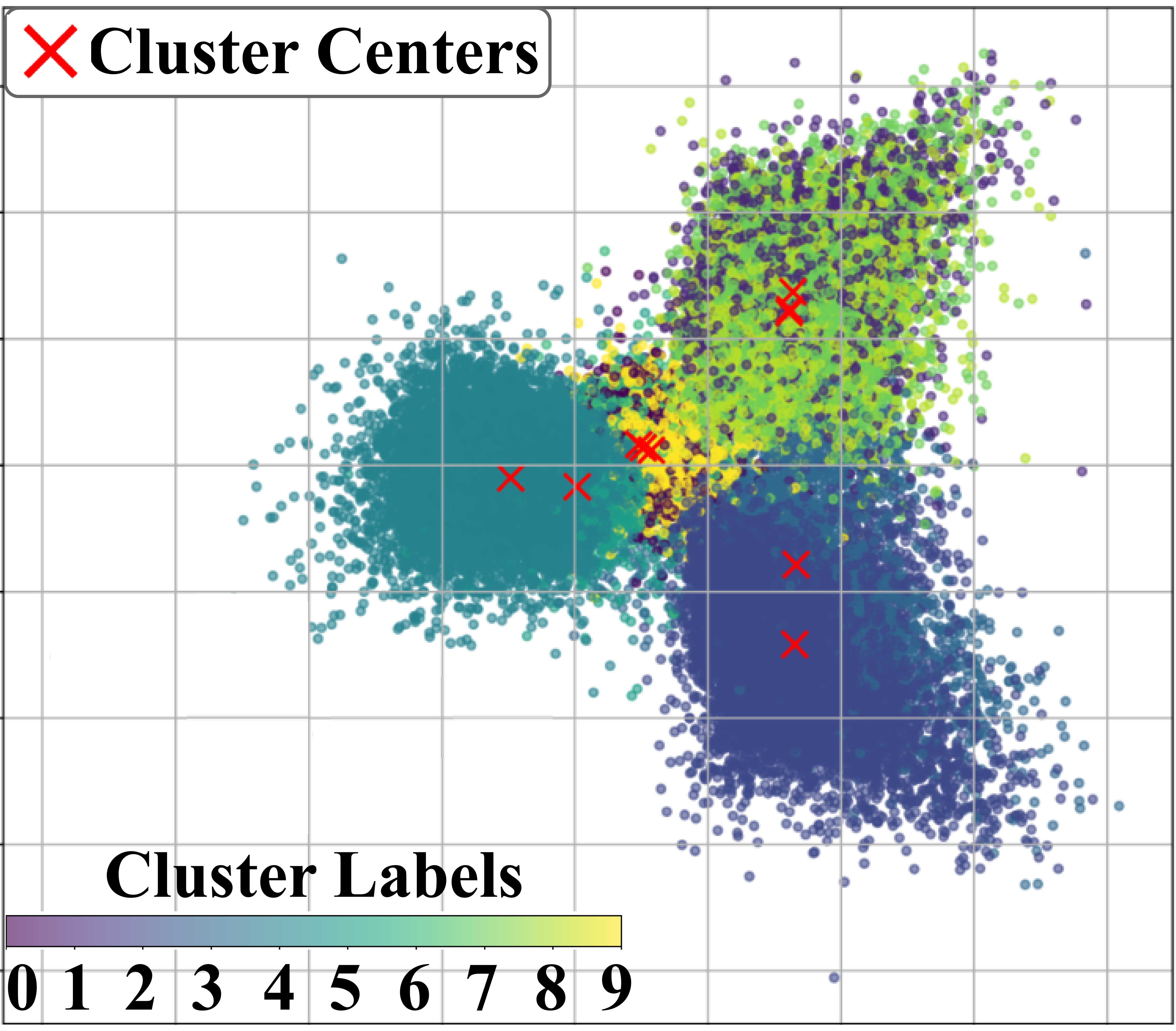}}
    \vspace{0cm} 
    \caption{Distribution of Item Attributes with and without Constraints.}
    \label{fig4}
\end{figure}

Additionally, we validate the effectiveness of the constraint in Equation (\ref{lcons}). The purpose of our constraint is to ensure that, after training with the backdoor attack, the distribution of candidate items remains stable, preventing the system from detecting the attack due to the inability to recommend items that should have been recommended to the target user before the attack. Figure \ref{fig4} (a) shows the distribution of candidate items after training without the constraint loss term, while (b) illustrates the feature distribution after incorporating the constraint. It can be observed that with the inclusion of the constraint loss, the item representations transition from a discrete state to one that aligns better with the distribution of each item category. Furthermore, Table \ref{table4} provides a more intuitive comparison of the fluctuation ratios of the candidate items for the target user with and without the constraint loss.

\begin{table}[tbp]
\centering
\renewcommand{\arraystretch}{1.5}  
\resizebox{0.48\textwidth}{!}{  
\begin{tabular}{c c c c c}
\toprule
\textbf{Datasets} & \textbf{Yelp} & \textbf{Gowalla} & \textbf{Amazon} & \textbf{MovieLens} \\ \hline
\textbf{Raw} & 0.1179 & 0.0769 & 0.0425 & 0.1487 \\
\textbf{w/o $L_{limit}$} & 0.0625 & 0.0469 & 0.0217 & 0.1162 \\
\hline
\textbf{Ours} & 0.0995 & 0.0715 & 0.0403 & 0.1369 \\
\bottomrule
\end{tabular}
}
\caption{Comparison of target item exposure in the clean graph, without constrained loss functions, and after applying our attack method.}
\label{table4}
\end{table}
We find that the model without the constraint term performs significantly worse in recommending candidate items even before the attack is launched, compared to the model with the constraint. This observation highlights that the constraint not only contributes to the effectiveness of the attack but also plays a crucial role in preserving the model’s recommendation quality under normal conditions. The performance drop in the unconstrained setting further validates the necessity of this design, as it ensures that the introduced triggers remain stealthy and do not degrade the overall user experience, thereby making the backdoor less detectable.

\begin{figure}[tbp]
    \centering
    \setcounter{subfigure}{0}
    \subfloat[Yelp]{
    \includegraphics[width=0.48\linewidth]{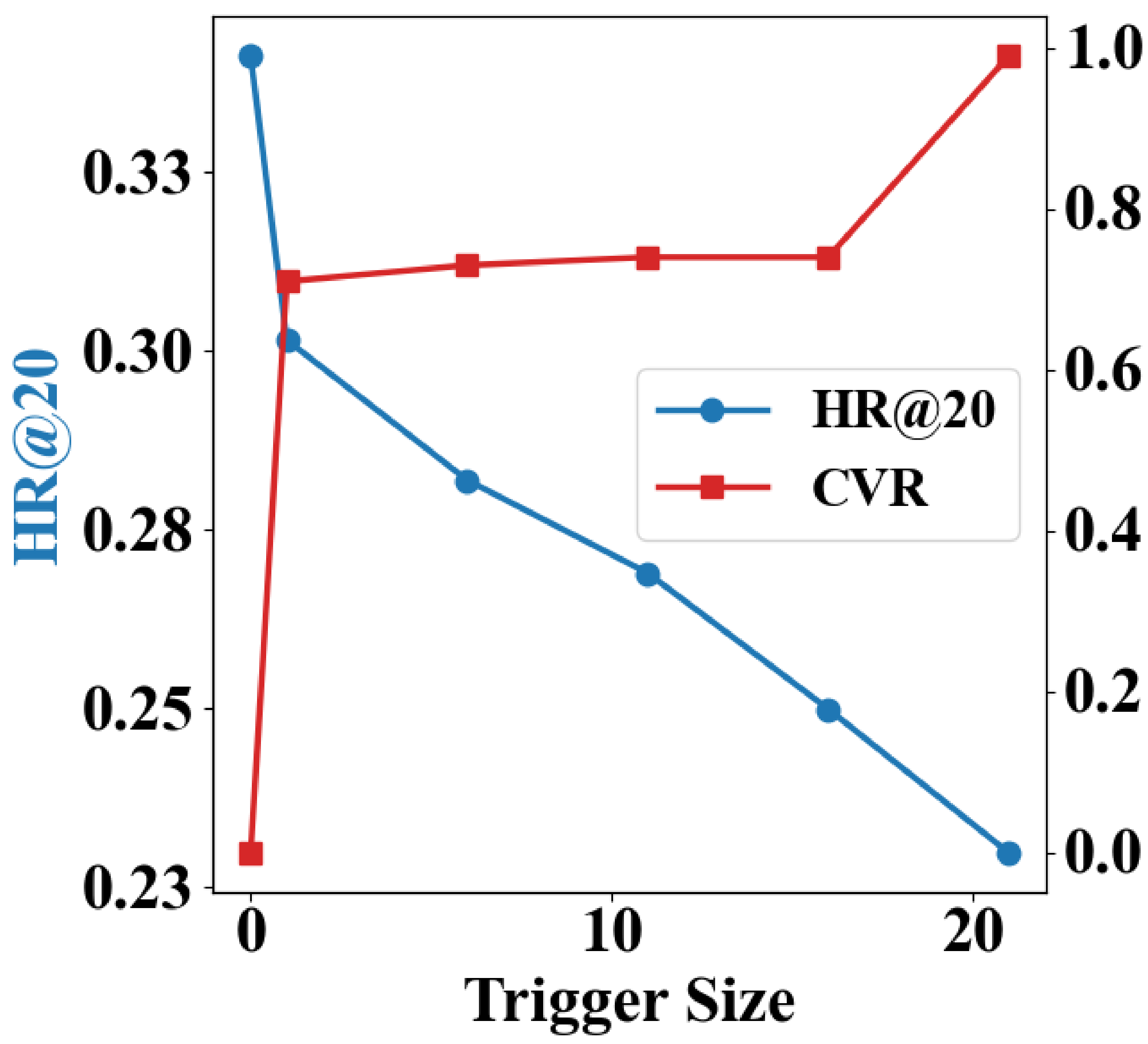}}
    \hspace{0cm}
    \subfloat[Gowalla]{
    \includegraphics[width=0.48\linewidth]{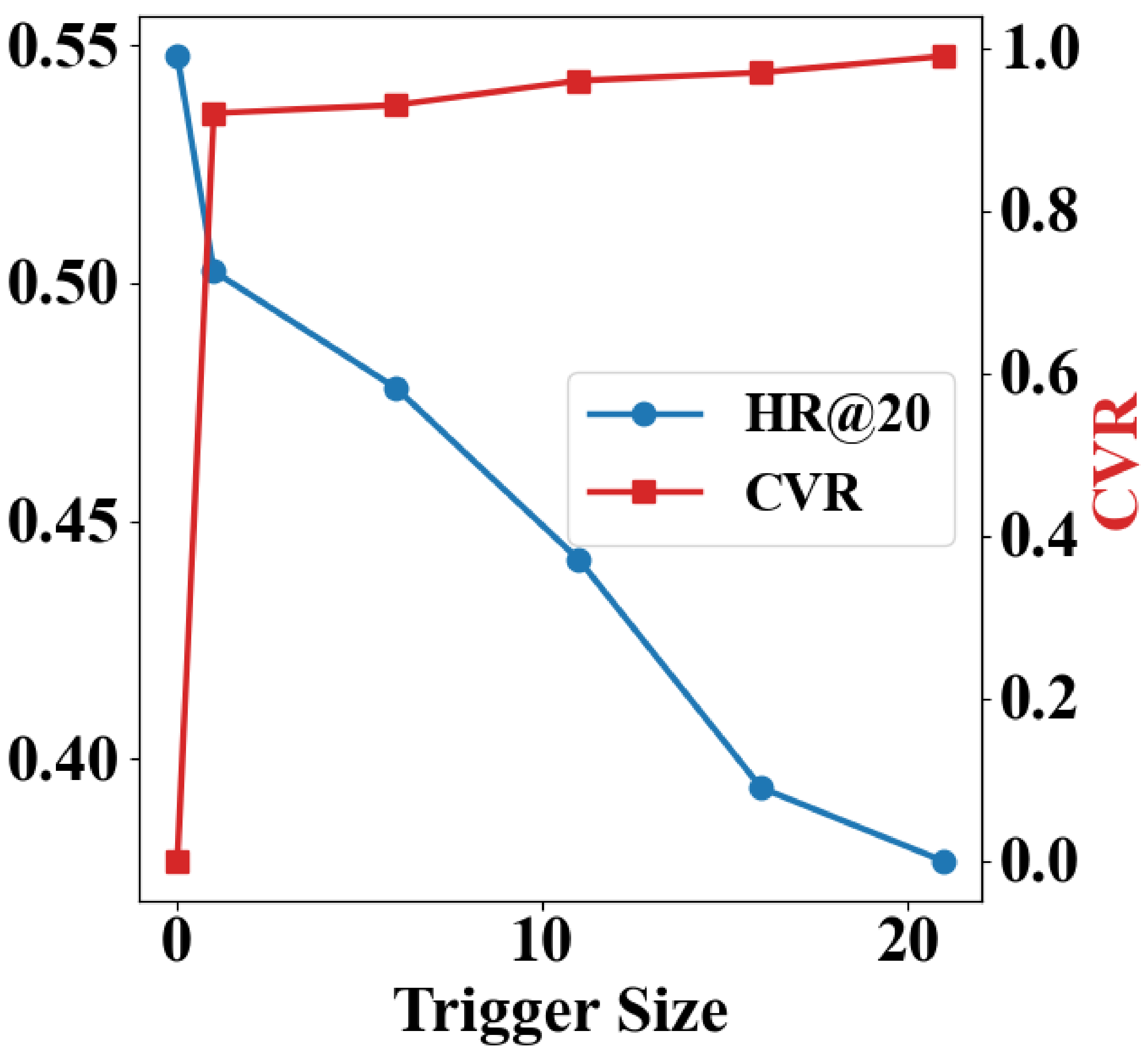}}
    \vspace{0cm} 
    \caption{We conduct comparative experiments on different sizes of triggers across four datasets.}
    \label{fig5}
\end{figure}

\noindent\textbf{Low Destructiveness:}
To answer \textbf{RQ3}, we will investigate the impact of multiple triggers on the experimental results. Specifically, we generate a trigger exclusively for a single target item. To ensure the authenticity of the trigger, its structure consists of the target item being connected to a fake user, which in turn is connected to a fake item. For each trigger, we set different control groups with varying numbers of fake items: $0$, $5$, $11$, $15$, and $20$. The attack results are shown in Figure \ref{fig5}.

We found that, although increasing the size of the trigger slightly improved the coverage of the target item, the extent of this improvement is negligible in practical applications. Additionally, as the size of the trigger increases, a large amount of fake information is propagated through the message-passing mechanism into the graph. Since the recommendation system is based on a bipartite graph structure and the proxy model generally uses three layers of convolution, the fake information can easily contaminate unrelated users and items, leading to significant performance degradation in the recommendation system. Even generating a trigger of size 1 for each target item individually can still cause a substantial decrease in performance.

Regarding this issue, we believe that when multiple triggers are present, they may overlap in their effects, especially when each trigger affects different target items. This can lead to redundant effects or conflicts, increasing the complexity of the recommendation system. The single trigger method, by uniformly linking all target items, avoids this redundancy and simplifies the attack structure. A single trigger can influence multiple target items, but its propagation path is controlled, preventing the complexity and unnecessary conflicts that may arise from the simultaneous use of multiple triggers.

\section{CONCLUSION}
In this paper, we propose a novel backdoor attack method for graph-based recommendation systems. This method generates a single trigger based on the target item, significantly increasing its exposure among the target users. Additionally, we design a constrained loss function to address the stealth issues inherent in traditional attack methods. By employing the single-trigger attack strategy, we mitigate the high destructiveness of false information on irrelevant user and item recommendations.
Our work also demonstrates that current mainstream graph recommendation methods lack robustness and are highly susceptible to various attacks. In future work, we will explore the detection of backdoor attacks in graph-based recommendation systems and investigate more robust recommendation models to fundamentally address the vulnerability to attacks.

\section*{Acknowledgments}
This work was partly supported by the National Natural Science Foundation of China (Nos. 92370111, 62272340, 62422210, 62276187, 62302333, U22B2036, 62261136549),  the Open Research Fund from Guangdong Laboratory of Artificial Intelligence and Digital Economy (SZ) (No. GML-KF-24-16), the Technological Innovation Team of Shaanxi Province (No. 2025RS-CXTD-009), the International Cooperation Project of Shaanxi Province (No. 2025GH-YBXM-017) and the Tencent Foundation and XPLORER PRIZE.


\bibliographystyle{named}
\bibliography{ijcai25}

\end{document}